# Automated Objective Surgical Skill Assessment in the Operating Room Using Unstructured Tool Motion

Piyush Poddar[1], Narges Ahmidi[2], S. Swaroop Vedula[2], Lisa Ishii[3], Gregory D. Hager[2], Masaru Ishii[3]

[1]Department of Biomedical Engineering, Johns Hopkins University, Baltimore, MD 21218,
ppod1991@gmail.com
[2]Department of Computer Science, Johns Hopkins University, Baltimore, MD 21218,
nahmidi1@jhu.edu  hager@cs.jhu.edu
[3]Department of Head and Neck Surgery-Otolaryngology, Johns Hopkins Medical Institutions,
Baltimore, MD 21287, {learnes2, mishii3}@jhmi.edu

**Abstract.** Previous work on surgical skill assessment using intraoperative tool motion in the operating room (OR) has focused on highly-structured surgical tasks such as cholecystectomy. Further, these methods only considered generic motion metrics such as time and number of movements, which are of limited instructive value. In this paper, we developed and evaluated an automated approach to the surgical skill assessment of nasal septoplasty in the OR. The obstructed field of view and highly unstructured nature of septoplasty precludes trainees from efficiently learning the procedure. We propose a descriptive structure of septoplasty consisting of two types of activity: (1) brushing activity directed away from the septum plane characterizing the consistency of the surgeon's wrist motion and (2) activity along the septal plane characterizing the surgeon's coverage pattern. We derived features related to these two activity types that classify a surgeon's level of training with an average accuracy of about 72%. The features we developed provide surgeons with personalized, actionable feedback regarding their tool motion.

## 1 Introduction

Academic surgical training programs typically rely on subjective approaches to evaluate trainees' surgical technical skill and to provide feedback to trainees [1,2]. In this paper, we study septoplasty — the surgical correction of a deviated nasal septum — which is an "index" surgery for residents in academic otolaryngology training programs. That is, trainees are expected to attain a certain level of proficiency with the procedure before completing graduate surgical training. Residents acquire proficiency with septoplasty by operating on patients under the supervision of an attending surgeon. Mucosal flap elevation is the most critical and technically challenging part of nasal septoplasty. During this portion of the procedure, surgeons use a Cottle elevator, which has a flat, spoon-shaped tip on one end, to break adhesions and elevate the mucosa from underlying septal cartilage and bone [3,4]. Septal mucosal flap elevation is an example of the broader category of unstructured dissection procedures, including laparoscopic dissections.

Evaluation of trainee performance by attending surgeons for septoplasty is challenging for several reasons. First, unlike many surgical procedures that can be decomposed into meaningful segments for teaching and feedback, mucosal flap elevation does not have a well-defined sequential structure. Second, attending surgeons cannot visualize the motion of the tool as trainees operate on the septum inside the nose. Consequently, attending surgeons cannot reliably assess skill and provide feedback regarding the trainee's choice of actions or on the trainee's efficiency and precision of tool motion. Additional challenges associated with automatic tracking of tool motion during surgery in patients include patient-specific anatomic variability, noise in sensor readings, patient movement, and multiple operating surgeons in a surgery [1].

Previous work on assessing surgical skill [5-9] using tool motion in the operating room (OR) has focused on structured surgical procedures such as cholecystectomy, and has used generic aggregate metrics such as path length, time, and the number of movements [8,9]. Since surgeons-in-training often repeatedly switch with attending surgeons mid-procedures, generic metrics become less meaningful for skill assessment and feedback. Further, generic metrics do not describe tool motion in a way that informs trainees on how to improve their future performance [10]. On the other hand, other statistical approaches such as Hidden Markov Models [11, 12] or Sparse-HMM [13] were successfully used in surgical skill assessment. It is yet ambiguous how a statistical model can be used for providing detailed and targeted feedback on a specific wrong surgical movement.

In our work, we aim to develop an automated method to assess surgical skill in an unstructured procedure using data from the operating room. To do so, we automatically detect strokes in tool motion, define two types of activity accomplished by the surgeons' strokes which characterize tool motion during septoplasty, and evaluate the discriminative ability of features representing these two types of activity to distinguish between expert and novice surgeons (Fig. 1).

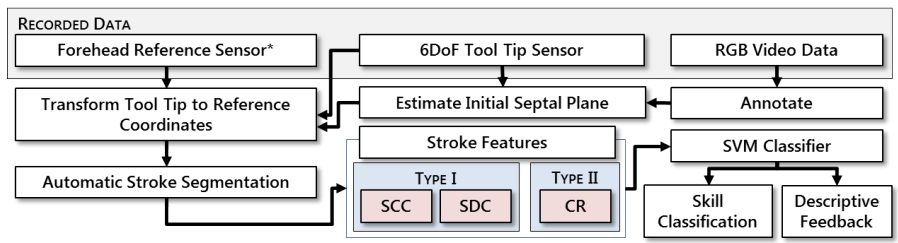

**Fig. 1.** System components and data flow for automatic skill classification in septoplasty- DoF: Degrees of Freedom, RGB: Red-Green-Blue, SCC: Stroke Curvature Consistency, SDC: Stroke Duration Consistency, CR: Coverage Rate, *For cases using a head sensor

## 2   Experimental Setup

The cases reported in this paper are the result of a substantial data collection effort over the course of one year, with roughly weekly data collection from four hospital sites participating in this study. We collected data from four expert surgeons (attending surgeons and surgical fellows) and seven novice surgeons (resident surgeons) as they performed mucosal flap elevation in septoplasty on patients. Our data was collected after obtaining Institutional Review Board approval and data capture did not in any way interfere with the care provided to patients. As shown in Fig. 2, an electromagnetic sensor (six degrees of freedom) was affixed to the Cottle elevator ("Cottle sensor"). The tool, along with the attached sensor, was sterilized using standard operating room procedures before each case. We tracked the Cottle sensor as surgeons elevated the mucosal flap during septoplasty using an electromagnetic field generator (Aurora®, Northern Digital, Inc., Ontario, Canada). Prior to the procedure, we performed a pivot calibration for both tips of the Cottle elevator. We recorded video of the procedure with two Kinects® (Microsoft, Inc., Redmond, WA) rigidly fixed to a tripod. For our analysis, we used data from 48 septoplasty cases, 28 of which were performed entirely by expert surgeons, 6 entirely by residents, and 14 by multiple operators. The multi operator cases were manually segmented to portions performed entirely by expert or residents, resulted a total of 42 expert and 20 novice trials for data analysis.

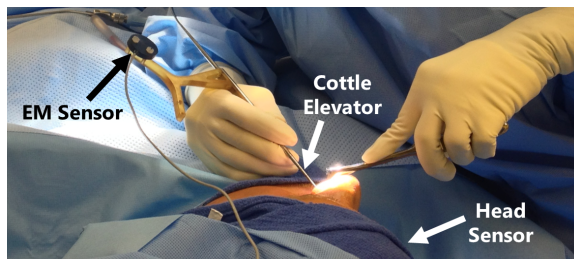

**Fig. 2.** Cottle elevator with affixed electromagnetic (EM) sensor used by surgeon to elevate mucosal flap in a septoplasty procedure. Head sensor (not visible) affixed to forehead region for the most recent third of cases to track head motion.

For the most recent third of the cases, we measured patient head motion using a second sterilized reference sensor attached to the sterile towel wrapped around the patient's head ("head sensor"). For cases where we did not use a reference sensor, we measured head motion by estimating the location of the septal plane from tool motion data at each frame. We assumed the Cottle elevator predominantly moved on the septal plane and that head motion was constrained to one degree of freedom (rotation around the neck). We estimated the septal plane at each frame to be the rotated initial septal plane that best fit the most recent tool tip motion data. Our estimation was cross-validated against the reference sensor data to ensure accuracy. The estimated plane was on average 3.4 degrees and 7mm off the real plane introducing ~6% error to the feature vector.

We transformed data captured from the Cottle sensor to the coordinates of the head pose, using either the reference sensor's coordinate system or the estimated head movements (Fig. 1).

At the beginning of each case, the surgeon was asked to move the Cottle tip around the perimeter of the nose to register the tool tip with the location of the patient's nose (Fig. 3 Left). No additional input from the surgeon was necessary for data capture. We estimated the initial septal plane as the plane formed by the first and third principal components of the active tool tip trajectory during nose registration.

We developed customized data collection software to record synchronized video and kinematic data. We used the video data to annotate time-points when surgeons initiated a different part of the procedure, switched between ends of the Cottle, switched with another surgeon, or started use or disuse of the Cottle elevator.

## 3 Methodology

In this section, we describe the automatic detection of strokes, development of features related to activities accomplished by the strokes, and classification of surgical skill using these features.

### 3.1 Automated Segmentation of Strokes

Through consultation with expert surgeons and exploratory data analyses, we see that surgeons elevate the mucosal flap during septoplasty by use of stroking motions of the Cottle elevator (Fig. 3 Left). To automatically extract such strokes, we defined the beginning of a stroke as the frame at which the Euclidean distance from the active tool tip to the septal plane is at a local minimum. We defined the end of a stroke as the closest frame following the start frame at which the distance from the active tool tip to the septal plane is a local maximum. To limit extraneous detections of strokes, we smoothed the tool tip position using a moving average filter and constrained the stroke time duration, length, and start/end distances to the center of the nose. We used this definition to automatically segment the motion of the Cottle elevator into discrete strokes.

### 3.2 Type I and Type II Activities

Each stroking motion during septoplasty is meant to accomplish two types of activities. Type I activity involves elevating the mucosal flap away from the septal cartilage and bone. Type I activity can be thought of as a brushing motion with the Cottle elevator and is related to the force applied by the surgeon to break adhesions between the mucosa and the septum and to elevate the mucosal flap (Fig. 3 Left). Type II activity reveals the surgeon's search pattern, i.e., how a surgeon covers the area of the septum while searching for adhesions between the mucosa and the cartilage. Type II activities may be represented by a 2D graph (Fig. 3 Right) that connects the starting positions of consecutive strokes after they have been projected onto the estimated septal plane.

The size of each vertex in the search graph shows the length of the corresponding stroke made at that point on the septum. The convex hull around the graph represents the cumulative area of the septum that has been elevated.

The distinction between Type I and Type II activities is clinically meaningful. Excessive force applied to elevate the mucosa (Type I activity) may tear the mucosa and result in septal perforation, which leads to undesirable postoperative outcomes. The extent and rate of septal plane coverage (Type II activity) reflects the surgeon's efficiency in elevating the mucosal flap.

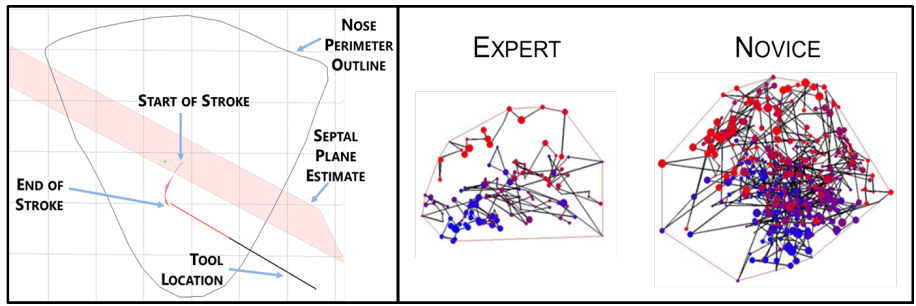

**Fig. 3.** (Left) Type I Activity: 3D Visualization of a detected stroke brushing away from septal plane, (Right) Type II Activity: Expert and novice 2D search graphs on the septal plane. Color (blue to red) indicates progression of time. The vertex size is proportional the length of a stroke. The red outline marks the convex hull representing the area of septum covered by the surgeon.

### 3.3 Feature Extraction

We specified three features based on a hypothesis that experts repeat strokes more consistently and regularly, and are more efficient relative to novice surgeons: (1) Stroke Curvature Consistency (SCC), (2) Stroke Duration Consistency (SDC), and (3) Coverage Rate (CR). We explored several additional features that could be extracted from the tool tip motion data, including stroke length, velocity, dominant frequencies, acceleration of the Type I activities, and variability in coverage rate as well as the distribution of angles, direction of edges, density of the vertices on the search graph, however the distribution of these features appeared to be similar between expert and novice trials and thus were not included in subsequent analyses.

**Stroke Curvature Consistency.**
Since different segments of septoplasty require different stroke types, measures of local variance rather than the standard global variance needed to be computed. SCC measures the local consistency of stroke curvatures across strokes. A consistent motion will yield strokes with similar curvatures and thus we expect experts to demonstrate a lower SCC than novice surgeons. We computed the curvature of the $i^{th}$ stroke, $c(i)$, to be the ratio of the stroke trajectory length to the Euclidean distance between the start and end points of the stroke. Using the vector of all curvatures **C**, we then measured the local consistency of curvatures by computing the median squared distance between C and its smoothed representation, $\hat{C}$:

$$SCC = median\left((C - \hat{C})^2\right), \quad \hat{C} = median\ filter(C) \tag{1}$$

**Stroke Duration Consistency.**

Similarly, SDC measures the local consistency in stroke duration:

$$SDC = median\left((\Delta t - \widehat{\Delta t})^2\right), \quad \widehat{\Delta t} = median\ filter(\Delta t) \tag{2}$$

where $\Delta t$ is the vector of all stroke durations and $\widehat{\Delta t}$ is the vector of all stroke durations after application of a median-filter. Locally consistent stroke durations, hypothesized to be practiced by experts, will result in a low SDC whereas locally variable stroke durations, seen in novices, will result in a high SDC.

**Coverage Rate.**
CR measures the rate of flap elevation and provides insight into how a surgeon covers the septum while searching for adhesions between the mucosal layer and the underlying septum. We defined $AC(i)$ as the area inside a convex hull of the search graph after completion of the $i^{th}$ stroke. The convex hull covers the finite set of points, $S(i)$, consisting of the vertex $s_1$ to vertex $s_i$ of the search graph.

$$AC(i) = Convex\ Area(S(i)); \quad S(i) = \{s_1, s_2, \dots, s_i\}; \quad i \geq 3 \tag{3}$$

We defined the CR to be the median increase in AC with each stroke:

$$CR = median(\dot{AC}); \quad \dot{AC}(i) = AC(i) - AC(i-1)\ for\ i = 4\ to\ N \tag{4}$$

where N is the number of vertices in the search graph. We hypothesized that the CR will be larger for experts compared with novice surgeons because experts elevate large areas of the mucosal flap with each stroke. In contrast, novice surgeons may fail to elevate the mucosal flap around adhesions and thus, they must re-elevate previously explored portions of the septum.

### 3.4 Automatic Skill Assessment

Exploratory analyses showed differences in the distributions of SDC, SCC, and CR across classes of surgical experience. We used a kernel support vector machine (SVM) classifier using SDC, SCC, and CR as the input feature vector and assigned ground-truth labels based on the surgeon's experience – attending surgeons and fellows as experts and residents as novices. We trained and tested the SVM classifier under two setups: leave-one-trial-out (TO) and leave-one-user-out (UO). In the TO setup, we used data from one trial (at a time) as the test data and data from the remaining trials for training. In the UO setup, we used all trials performed by one surgeon as the test data and data from the remaining surgeons for training. Trials per-formed by two operators were considered to be two separate trials with all data from a single operator concatenated together. Trial-level features were computed as the median of 'sub-trial level' features. That is, within a trial, each feature was computed once for each continuous instance of cottle use i.e. each sub-trial. The resulting trial-level feature was the median of these sub-trial feature computations. We excluded from our analysis all trials where fewer than seven strokes were performed in each sub-trial. We computed the micro-average (ratio of correctly classified samples to the total number of samples) and macro-average accuracy (average of true positive rates of each class) of our classifier as a measure of its ability to discriminate surgical skill.

## 4  Results and Discussions

Our exploratory analyses revealed that the distribution for SCC, SDC, and CR appeared to be different for expert and novice trials. We observed that experts demonstrated decreased SCC relative to novices suggesting a more consistent brushing motion. We observed, contrary to our hypothesis, a higher SDC for experts than for novices, indicating increased local stroke duration variability. The observed variability in SDC may be because experts were able to better adapt their technique to differences in patient anatomy. Expert surgeons covered the area of the septum more quickly than novice surgeons resulting in a higher CR (Fig. 4). That is, experts elevated larger areas of the mucosal flap per stroke. Our classifier discriminated between trials performed by expert and novice surgeons under both the TO and UO setups with an overall micro-average accuracy of 69.6% for both TO and UO and overall macro-average accuracy of 72.1% for TO and 73.3% for UO (Tables 1 and 2). The classification accuracy using each individual feature was similar to the accuracy obtained using all three features.

We also modeled the expertise level using Hidden Markov Model (HMM) approach with 3 states and one mixture per state, with the same configuration reported in [11, 12]. The HMM model achieved 78.77% accuracy for UO and 54.89% for TO, whereas our method (with 73% and 72% accuracy) has more uniform performance on unseen users and can provide skill level information for specific types of surgeon motions.

## 5  Conclusion

In this paper, we developed and validated an objective method of assessing surgical skill and describe unstructured tool motion during a surgical procedure using features that are useful for descriptive, actionable feedback. The features, developed based on expert surgeons' understanding of the surgical procedure, are meaningful for providing feedback to trainees. For example, a higher-level feedback may be focused on efficient strategies to search for adhesions between the mucosal flap and the underlying septal cartilage, or on practicing the stroke motion such that trainees learn to elevate larger areas using optimal force or to adapt to changes in patient anatomy. Some possible lower-level feedbacks that our method can provide for trainees are for exam-ple: a trainee should practice stronger strokes to elevate wider areas of the mucosal flap (Coverage Rate feature); this strong stroke however should be performed with a curved and consistent wrist motion (Stroke Curvature Consistency feature) and with the same speed as the previous stroke (Stroke Duration feature).

**Table 1.** Classification performance (%) using leave-one-trial-out (TO) setup (chance=50%). E= Expert, N=Novice, GT= Ground Truth

| Only SCC | | GT E | GT N | Only SDC | | GT E | GT N | Only CR | | GT E | GT N | Overall (SCC, SDC, CR) | | GT E | GT N |
|---|---|---|---|---|---|---|---|---|---|---|---|---|---|---|---|
| Pre dict | E | 76.3 | 55.6 | Pre dict | E | 50.0 | 5.6 | Pre dict | E | 60.5 | 22.2 | Pre dict | E | 55.3 | 11.1 |
| | N | 23.7 | 44.4 | | N | 50.0 | 94.4 | | N | 39.4 | 77.8 | | N | 44.7 | 88.9 |
| MicroAvg | | 66.1 | | MicroAvg | | 69.6 | | MicroAvg | | 66.1 | | MicroAvg | | 69.6 | |
| MacroAvg | | 60.4 | | MacroAvg | | 72.2 | | MacroAvg | | 69.2 | | MacroAvg | | 72.1 | |

**Table 2.** Classification performance (%) using leave-one-user-out (UO) setup (chance=50%), E= Expert, N=Novice, GT= Ground Truth

| Only SCC | | GT E | GT N | Only SDC | | GT E | GT N | Only CR | | GT E | GT N | Overall (SCC, SDC, CR) | | GT E | GT N |
|---|---|---|---|---|---|---|---|---|---|---|---|---|---|---|---|
| Pre dict | E | 89.5 | 55.6 | Pre dict | E | 57.9 | 5.6 | Pre dict | E | 68.8 | 22.2 | Pre dict | E | 63.2 | 16.7 |
| | N | 10.5 | 44.4 | | N | 42.1 | 94.4 | | N | 34.2 | 77.8 | | N | 36.8 | 83.3 |
| MicroAvg | | 75.0 | | MicroAvg | | 69.6 | | MicroAvg | | 69.6 | | MicroAvg | | 69.6 | |
| MacroAvg | | 67.0 | | MacroAvg | | 76.2 | | MacroAvg | | 71.8 | | MacroAvg | | 73.3 | |

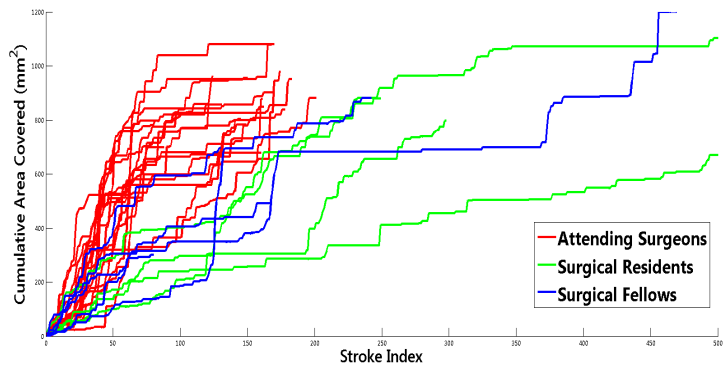

**Fig. 4.** Cumulative areas for single-operator trials performed by attending and trainee surgeons; CR is the median slope of each curve. Attending surgeons are considered more skilled than fellows and fellows better than residents in training.

Although our work is focused on nasal septoplasty, our features may be applicable to other segments of surgical procedures that are characterized by unstructured tool motion such as blunt dissection. Supplementing our data with additional surgeons and with additional features, including generic metrics of motion efficiency may improve the ability of a classifier to discriminate between levels of surgical skill. This study was limited by only having self-proclaimed skill level as the ground truth. Using a case-by-case rating system for measuring the ground-truth skill level may be more appropriate as some novice surgeons may perform in a manner similar to expert surgeons. Future work includes using a subjective grading-system as the ground truth and to evaluate the validity of providing real-time feedback regarding the developed features and its effect on trainee skill acquisition.

**Acknowledgment**. We gratefully acknowledge support from NIH 5R21DE022656-02. We would also like to thank participating attending surgeons - Drs. Kofi Boahene, and Patrick Byrne and trainee surgeons - Drs. Sun Ahn, Amit Kocchar, Linda Lee, Ryan Li, Myriam Loyo, Sofia Lyford-Pike, Peter Revenaugh, David Smith, and Babar Sultan at the Johns Hopkins Medical Institutions.